\pdfoutput=1
\documentclass[11pt]{article}

\usepackage[]{ACL2023} % setelah diedit jadi hilang num-pagenya

\usepackage{times}
\usepackage{latexsym}

\usepackage[T1]{fontenc}
\usepackage[utf8]{inputenc}

\usepackage{microtype}

\usepackage{inconsolata}

\usepackage{graphicx}
\usepackage{caption}
\usepackage{subcaption}

\usepackage{amsmath}
\usepackage{bbm}
\usepackage{graphicx}
\usepackage{tikz}
\usepackage{hyperref}
\usepackage{colortbl}

\usepackage{multirow}
\usepackage{booktabs}
\usepackage{amssymb}
\usepackage{makecell}

\title{IndoToD: A Multi-Domain Indonesian Benchmark For End-to-End Task-Oriented Dialogue Systems}

\author{Muhammad Dehan Al Kautsar$^1$, Rahmah Khoirussyifa' Nurdini$^1$, Samuel Cahyawijaya$^3$, \\ {\bf Genta Indra Winata$^2$, Ayu Purwarianti$^1$} \\
$^1$Institut Teknologi Bandung\quad $^2$Bloomberg
\\ $^3$The Hong Kong University of Science and Technology \\
\texttt{\href{mailto:dehanalkautsar3@gmail.com} {\color{black}{dehanalkautsar3@gmail.com}}}
}

\begin{document}
\maketitle
\begin{abstract}
Task-oriented dialogue (ToD) systems have been mostly created for high-resource languages, such as English and Chinese. However, there is a need to develop ToD systems for other regional or local languages to broaden their ability to comprehend the dialogue contexts in various languages. This paper introduces IndoToD, an end-to-end multi-domain ToD benchmark in Indonesian. We extend two English ToD datasets to Indonesian, comprising four different domains by delexicalization to efficiently reduce the size of annotations. To ensure a high-quality data collection, we hire native speakers to manually translate the dialogues. Along with the original English datasets, these new Indonesian datasets serve as an effective benchmark for evaluating Indonesian and English ToD systems as well as exploring the potential benefits of cross-lingual and bilingual transfer learning approaches.
\end{abstract}

\section{Introduction}

% Short paper 1 halaman, long paper biasanya 1.5 halaman. Bisa liat contoh introductionnya BiToD/CamRest. 

\begin{figure*}[!t]
\centering
\resizebox{\linewidth}{!}{
\includegraphics{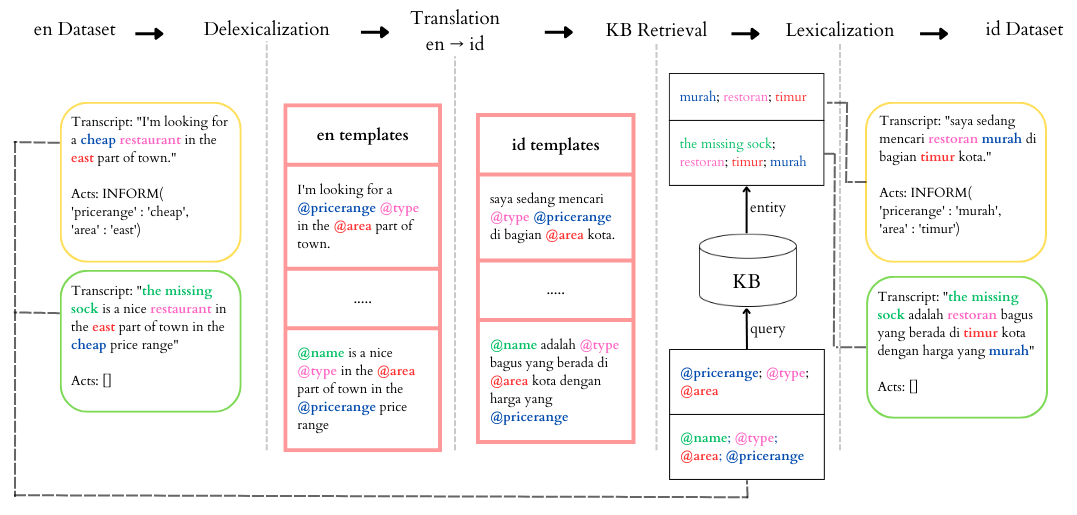}
}
\caption{Illustration of the dialogues collection pipeline: a) Delexicalize dialogue conversations to create English templates; b) Translate English conversation templates to Indonesian;
c) Retrieve KB to retrieve corresponding entities; d) Lexicalize the dialogue conversations to collect Indonesian dialogues.}
\label{fig:datasetcollection}
\end{figure*}

% paragraf 1 jelasin ToD dan its difficulties, jelasin knp harus ada multilingual dialogue system
Task-oriented dialogue (ToD) systems are conversational agents designed to interact with users and assist them in various domains, such as restaurant~\cite{bordes2016learning,wen2017network}, public transport~\cite{budzianowski2018multiwoz,lin2021bitod}, and in-car assistance~\cite{eric2017key}. This system also serves as the base for many commercial products that operate using the dialogue systems approach because of its ability to operate and understand the dialogue context without using hand-crafted rules~\cite{lin2021bitod}. 
Despite the recent growing interest in developing end-to-end ToD systems due to their simplicity, ToD systems are mostly created using monolingual datasets in high-resource languages such as English and Chinese. 
% particularly in the end-to-end approach because of its simplicity over the modular approach, 
Moreover, building a ToD system can be challenging due to the limited availability of datasets for training and evaluating the system, which has been identified as the most critical factor in preventing the creation of bilingual/multilingual ToD systems~\cite{wen2017network, Razumovskaia2021CrossingTC}. 

% As such, 
Developing ToD systems in additional underrepresented languages is essential to expand their capabilities to understand dialogue contexts in diverse languages~\cite{kanakagiri2021task}. One of them is Indonesian, a language spoken by many people worldwide yet still considered an underrepresented language in the end-to-end ToD system.
% paragraf 2 jelasin ke b.indo yg underrepresented in ToD
Indonesian is ranked as the fourth most internet users in the world based on the latest data per country~\cite{aji-etal-2022-one}, with around 212 million internet users.\footnote{\url{https://www.internetworldstats.com/stats3.htm}} However, the language itself is still categorized as an underrepresented language in the natural language processing (NLP) community because of problems such as scattered datasets and minimum community engagement~\cite{wilie2020indonlu,cahyawijaya2021indonlg,cahyawijaya2022nusacrowd}. 
% Regarding the ToD task, 
To the best of our knowledge, there is only one publicly available end-to-end Indonesian ToD dataset\footnote{\url{https://indonlp.github.io/nusa-catalogue/}} which is COD \cite{majewska2023cross}, a multilingual ToD dataset that is solely used for evaluation -- only test set available. 
% end-to-end ToD dataset
It has a very limited number of samples for Indonesian with 194 dialogues across 11 domains ($\sim$18 dialogues per domain) and no training data provided.
% that has not focused in Indonesian. 
This emphasizes the need to create larger end-to-end Indonesian ToD datasets to expand the capabilities to build and evaluate localized Indonesian ToD systems.

% paragraf 3: oleh krn itu, kami membuat...
To address the aforementioned issues, we propose \textbf{IndoToD}, an end-to-end multi-domain ToD benchmark in Indonesian. IndoToD comprises the collection of two parallel Indonesian end-to-end ToD datasets covering four different domains by manually translating two existing English datasets: CamRest676~\cite{wen2017network} and SMD~\cite{eric2017key} using delexicalization and lexicalization processes, as well as the evaluation of existing end-to-end ToD frameworks in various settings such as monolingual, cross-lingual, and bilingual.
% paragraf 4: jelasin dataset ini bisa digunakan buat a,b,c.
IndoToD provides more dialogue samples compared to~\cite{majewska2023cross} that can be utilized for training and evaluation.

% either as training data, evaluation data, or both.
% And due to the process of creating two parallel datasets, 
% We compare the performance of the existing English and modified Indonesian ToD systems using several metrics that are used to evaluate ToD systems.

% paragraf 5: contribution of this work
This paper's contributions are summarized three-fold as follows: 
\begin{itemize}
    \item We introduce IndoToD, the multi-domain benchmark for Indonesian ToD systems. The benchmark comprises two datasets: IndoCamRest and IndoSMD, with four different domains that serve as resources for training and evaluation. 
    \item We establish baselines on monolingual, bilingual, and cross-lingual settings on existing end-to-end ToD frameworks.
    \item We analyze the effectiveness of training bilingual datasets to leverage more training data and handle tasks in both languages in building ToD systems.
    % compared to monolingual English and Indonesian on ToD systems.
\end{itemize}

\section{IndoToD Benchmark}

%%%%%%%%%%%%%%%%%%%
% table 1: comparison between COD (existing Indonesian dataset), IndoCamRest, and IndoSMD 
\begin{table*}
\centering
\resizebox{0.98\linewidth}{!}{
\begin{tabular}{p{.25\linewidth}p{.25\linewidth}p{.25\linewidth}p{.25\linewidth}} 
 \toprule
 \multicolumn{1}{c}{\textbf{Statistics}} & \multicolumn{1}{c}{\textbf{COD-id}} & \multicolumn{1}{c}{\textbf{IndoCamRest (ours)}} & \multicolumn{1}{c}{\textbf{IndoSMD (ours)}}   \\ \midrule
 \# domains & 11: \texttt{Flights, Weather, Alarm, Ride Sharing, Movies,  Music, Media, Homes, Banks, Travel, Payment} & 1: \texttt{Restaurant Search} & 3: \texttt{POI navigation, Calendar Scheduling, Weather Information} \\ \midrule
  \# dialogues & 194 & 676 & 323 \\ 
 \# dialogues/domain & 17.64 & 676 & 107.67 \\ 
 \# distinct entities & N/A & 110 & 325 \\ 
 \# turns/dialogue & 12.83 & 4.06 & 2.63 \\  
 \# tokens/utterance & N/A & 9.54 & 8.14 \\ 
 \# vocab (lexicalized) &  N/A & 1,103 & 883 \\ 
 \# vocab (delexicalized) &  N/A & 829 & 577 \\ \bottomrule
\end{tabular}
}
\caption{Comparison of our IndoToD benchmark with the Indonesian subset of COD~\cite{majewska2023cross}.}
\label{tab:datasetcomparison}
\end{table*}

IndoToD benchmark\footnote{\url{https://github.com/dehanalkautsar/IndoToD}} is created to develop the Indonesian ToD system. The benchmark covers four different domains, i.e., \texttt{restaurant search}, \texttt{point-of-interest (POI) navigation}, \texttt{calendar scheduling}, and \texttt{weather information}. IndoToD extends the existing English datasets, and we follow a streamlined process to conduct dialogue collection. The dialogues are multi-turn conversations that involve two speakers (i.e., user $U$ and system $S$). For each conversation, there is a knowledge base (KB) for the system to generate the correct entity for the user.

% We focus on creating the Indonesian ToD benchmark based on two existing English ToD datasets. 

\subsection{Dataset Collection}

We use CamRest676~\cite{wen2017network} and SMD~\cite{eric2017key} as the original English datasets that will go through several stages before becoming datasets that can be used in various Indonesian Task-oriented Dialogue (ToD) system experiments. CamRest676 is a ToD dataset that focuses on restaurant search queries collected via Wizard-of-Oz (WoZ) framework~\cite{Kelley1984AnID}. It consists of a collection of dialogues between the user and the system, where each dialogue has a task-specific goal (e.g., finding a restaurant). 
In this dataset, the user acts as a client who requests restaurant information and the system acts as an information provider that guides and answers the user's request. 

Meanwhile, SMD is an in-car assistant multi-domain ToD dataset. It covers several domains such as POI navigation, calendar scheduling, and weather information. This dataset was also created by using the same WoZ framework to get a high-quality dataset that imitates a conversation between two individuals in a way that resembles a real-life interaction between a driver and an in-car assistant. In the WoZ framework, the source of conversation is created through a human-to-human dialogue conversation which is collected through crowd-sourcing. Because of that, the conversations between the user and the system are more natural. 

% We utilize these datasets because of the collection methods that they used. 

% Moreover, these datasets' number of domains and dialogues are suitable for building datasets. 
% Both datasets have an adequate number of dialogues that are neither too large nor too small. A large dataset could impact the annotation process in creating the synthetic dataset, while a small dataset may not provide sufficient data for the training process. Then, the difference in the number of domains of the two datasets can provide diverse variations to build a challenging Indonesian ToD datasets.

These datasets then will be used as a reference for generating a ToD dataset that mimics natural conversation between two speakers and includes relevant knowledge base information for generating accurate responses. Both two datasets provide a diverse variety of dialogues since they have a different set of domains that complement the IndoToD benchmark.

\subsection{Dataset Construction}

The datasets have been created through several steps, as shown in Figure~\ref{fig:datasetcollection}. First, existing English datasets are delexicalized into several user and system template sentences using \citet{madotto2020learning} approach. Then, these template sentences are translated by native Indonesian annotators into Indonesian sentences. Lastly, new Indonesian datasets are built through the KB retrieval and lexicalization process.

\paragraph{Delexicalization} Initially, the datasets are pre-processed by delexicalizing the entities. The process is carried out using the source code implemented by \citet{madotto2020learning} to remove the entities in the dataset's dialogue. The aim of this step is to reduce the number of dialogues that require annotation (translation) by replacing all entities with a common, pre-defined value based on the type of entity. This pre-processing method is effective because the number of sentences that need to be translated is reduced by 30\% and 34\% for CamRest676 and SMD, respectively. The output of this step is a list of pre-processed, entity-less sentences for each dataset.

\paragraph{Translation} The list of pre-processed sentences is translated into Indonesian by annotators. In total, there are 3,834 and 1,126 sentences translated from CamRest676 and SMD, respectively. This process is followed by a cross-validation process where an annotator has to check the work of other annotators to maintain the dataset's quality. Note that for the SMD dataset, we only use 11\% (323 dialogues) of its original dialogues due to the limitations on the annotator side.

\paragraph{KB Retrieval and Lexicalization} After the translation process, 
% the Indonesian ToD datasets are ready to be generated. 
to build each dialogue, a collection of Indonesian sentences and their corresponding entities are needed to begin the lexicalization process. We retrieve the Indonesian entities by querying the KB for the required slots to construct the dialogue. The lexicalization process is conducted by using Indonesian entities and sentences that have been retrieved earlier. This process ends with an evaluation of 100 randomly selected newly created Indonesian dialogues by human evaluators to detect any errors to ensure the quality of the resulting dialogues. If there are any errors, we ask them to manually edit that utterance. The sample of corrected Indonesian utterances is presented in Appendix Table~\ref{tab:sampledevaluationfixing}.

%%%%%%%%%%%%%%%%%%%%%%%%%%%%
\begin{table*}
\centering
\resizebox{0.98\linewidth}{!}{
    \begin{tabular}{c|p{.2\linewidth}|p{.2\linewidth}p{.2\linewidth}p{.2\linewidth}}
    \toprule
    % \multirow{2}{*}{\textbf{Statistics}} & \multirow{2}{*}{\textbf{IndoCamRest (Restaurant Search)}} & \multicolumn{3}{c}{\textbf{IndoSMD}} \\
    \multirow{2}{*}{\textbf{Statistics}} & \multicolumn{1}{c|}{\textbf{IndoCamRest}} & \multicolumn{3}{c}{\textbf{IndoSMD}} \\
    \cmidrule{2-5} & \texttt{Restaurant Search} & \texttt{POI Navigation} & \texttt{Calendar Scheduling} & \texttt{Weather Information} \\ \midrule
    \# dialogues & 676 & 100 & 114 & 109 \\ 
    \# slot types & 6 (\texttt{area}, \texttt{food}, \texttt{price} \texttt{range}, \texttt{postcode}, \texttt{phone}, \texttt{address}) & 5 (\texttt{POI name}, \texttt{traffic info}, \texttt{POI category}, \texttt{address}, \texttt{distance}) & 6 (\texttt{event}, \texttt{time}, \texttt{date}, \texttt{party}, \texttt{room}, \texttt{agenda}) & 4 (\texttt{location}, \texttt{weekly time}, \texttt{temperature}, \texttt{weather attr.}) \\ 
    \# distinct slot values & 88 & 131 & 79 & 78 \\ \bottomrule
    \end{tabular}
}
\caption{Per-domain statistics of IndoCamRest and IndoSMD.}
\label{tab:perdomainstatistics}
\end{table*}
%%%%%%%%%%%%%%%%%%%%%%%%%%%%

\subsection{Dataset Statistics}

We collect a total of 999 dialogues with 6,338 utterances from two datasets, named as IndoCamRest and IndoSMD, which are derived from CamRest676 and SMD datasets, respectively.
We compare several statistic attributes between our datasets (IndoCamRest and IndoSMD) and COD that are shown in Table~\ref{tab:datasetcomparison}.
Furthermore, the per-domain statistics of IndoCamRest and IndoSMD are presented in Table~\ref{tab:perdomainstatistics}. IndoCamRest and IndoSMD share some of the same statistical characteristics as the previous datasets, including the number of domains, intents, and slot types. It is worth noting that both of our datasets have a substantially larger number of dialogues per domain compared to COD. In addition, IndoCamRest and IndoSMD comprise both the lexicalized and delexicalized forms in the dialogue. The inclusion of the delexicalized form is intended to facilitate the generation of dialogues by lexicalizing the delexicalized sentences. 
% IndoSMD has two versions such as a split set and a per-domain set, depending on the specific tasks that must be performed.

\section{Experimental Setup}
\subsection{Experiment Settings}
We set up a benchmark for both Indonesian and English ToD to evaluate the performance of the current ToD systems. We explore three different training and evaluation settings:

\begin{itemize}
    \item \textbf{Monolingual -- id/en}. We train the end-to-end ToD systems using monolingual corpus in each language independently.
    \item \textbf{Cross-lingual}. We train the systems using the English end-to-end ToD dataset before testing it using an Indonesian test set to analyze the effectiveness of the cross-lingual approach in understanding the context of the dialogue.
    \item \textbf{Bilingual -- id+en}. We train the systems by combining English and Indonesian datasets, utilizing parallel corpora from CamRest and SMD. This allows us to examine how the system's performance is affected by exposure to bilingual languages.
\end{itemize}

% , i.e., monolingual (using the same language corpus for the training and evaluation phase), cross-lingual (using English for training and Indonesian for the evaluation phase), and bilingual (using combined English and Indonesian training and test sets).

% The current performance of ToD systems in each language, English and Indonesian, is evaluated through the monolingual setting. The monolingual setting is conducted by using the same monolingual corpus in the training and test phase, both in Indonesian and English. Furthermore, the zero-shot cross-lingual learning setting is conducted by using different types of datasets in the training and testing phase. English corpus is used for the training phase and the Indonesian corpus for the test phase. This setting is conducted to evaluate the ToD system's contextual understanding capability, even when trained on a corpus of a different language.

% We evaluate the ToD system's performance in a bilingual setting, where it is trained on a corpus containing both English and Indonesian. This allows us to examine how the system's performance is affected by exposure to bilingual languages. In this setting, the ToD systems are being trained using Indonesian and English corpus, and then test them using Indonesian or English ToD datasets.

For monolingual and bilingual settings, we test each ToD framework in English and Indonesian test sets separately to analyze the impact of monolingual and bilingual training in each language. 
We also split the data into 3:1:1 and 8:1:1 for the CamRest and SMD, respectively.

%#### Table for results ####%
%## test on indonesian ##%
\begin{table*}
\centering
\footnotesize
\begin{tabular}{ l|cccc|cccc }
\toprule
\multirowcell{2}{\textbf{Baselines}} & \multicolumn{4}{c|}{\textbf{IndoCamRest}} & \multicolumn{4}{c}{\textbf{IndoSMD}} \\ 
\cmidrule{2-9} & \textbf{BLEU} & \textbf{Match} & \textbf{Success} & \textbf{Combined} & \textbf{BLEU} & \textbf{Match} & \textbf{Success} & \textbf{Combined} \\ \cmidrule{1-9}
\multicolumn{9}{c}{\texttt{monolingual}} \\ \cmidrule{1-9}
Sequicity \cite{lei2018sequicity} &16.74	&92.31	&81.29	&103.54	&\textbf{8.75}	&\textbf{44.83}	&\textbf{61.04}	&\textbf{61.69} \\
LABES \cite{zhang2020probabilistic} &17.59	&70.00	&78.49	&91.83	&1.01	&0.00	&34.82	&18.42 \\
MinTL \cite{lin2020mintl} &14.61	&80.33	&80.31	&94.93	&0.45	&24.14	&23.70	&24.37 \\
GALAXY \cite{He2021GALAXYAG} &\textbf{17.89}	&\textbf{96.24}	&\textbf{84.62}	&\textbf{108.32} &- &- &- &-\\ \cmidrule{1-9}
\multicolumn{9}{c}{\texttt{cross-lingual}} \\ \cmidrule{1-9}
Sequicity \cite{lei2018sequicity} &0.08	&8.46	&31.58	&20.10	&0.00	&15.52	&29.46	&22.50 \\
LABES \cite{zhang2020probabilistic} &0.02	&0.00	&60.96	&30.50	&0.00	&\textbf{15.79}	&55.84	&\textbf{35.82}\\
MinTL \cite{lin2020mintl} &\textbf{1.61}	&\textbf{16.67}	&\textbf{68.81}	&\textbf{44.35}	&0.00	&13.33	&\textbf{56.75}	&35.04 \\
GALAXY \cite{He2021GALAXYAG} &0.05	&0.00	&47.24	&23.67	&0.00	&0.00	&37.11	&18.56 \\ \cmidrule{1-9}
\multicolumn{9}{c}{\texttt{bilingual}} \\ \cmidrule{1-9}
Sequicity \cite{lei2018sequicity} &15.84	&89.23	&\textbf{84.01}	&102.46	&10.95	&\textbf{58.33}	&75.66	&\textbf{77.94} \\
LABES \cite{zhang2020probabilistic} &\textbf{18.77}	&\textbf{96.36}	&81.68	&\textbf{107.79}	&8.78	&47.37	&71.11	&68.02 \\
MinTL \cite{lin2020mintl} &18.65	&87.30	&79.63	&102.12	&\textbf{11.59}	&48.33	&\textbf{77.86}	&74.68 \\
GALAXY \cite{He2021GALAXYAG} &17.09	&94.74	&83.60	&106.26	&0.08	&45.00	&9.17	&27.17 \\
\bottomrule
\end{tabular}
\caption{Experiment settings result on Indonesian test set. \textbf{bold} denotes the best score per metric.}
\label{tab:indoresults}
\end{table*}

\subsection{ToD Baselines}
As baselines, We evaluate four ToD frameworks: Sequicity \cite{lei2018sequicity}, LABES \cite{zhang2020probabilistic}, MinTL \cite{lin2020mintl}, and GALAXY \cite{He2021GALAXYAG}. Since the four ToD system frameworks are designed for English, we have made adjustments to the frameworks to suit the requirements of our experiments. We largely adopt the configuration settings used in the original paper, including batch size, decoding method, early stop count, and learning rate.

To adapt the Sequicity and LABES frameworks for our experiments, we utilize fastText \cite{bojanowski2017enriching} instead of GloVe~\cite{pennington-etal-2014-glove}. This was necessary because fastText can accommodate English, Indonesian, and bilingual Indonesian-English word vectors \cite{conneau-etal-2017-muse}. Specifically, we incorporate common crawl\footnote{\url{https://fasttext.cc/docs/en/crawl-vectors.html}} English and Indonesian fastText word vectors for monolingual English and Indonesian experiment settings respectively. Meanwhile, we use bilingual English-Indonesian word vectors \cite{conneau-etal-2017-muse} for other settings.

Unlike the other frameworks, MinTL utilizes T5 \cite{raffel2020exploring} and BART \cite{lewis2020bart} as its backbone, which sets it apart regarding approach and potential performance. Note that for this experiment, we only use T5 \cite{raffel2020exploring} and mT5 \cite{xue2021mt5} as the backbone models, and use this framework without utilizing \emph{Levenshtein Belief Spans (Lev)} resulting in an inference process similar to Sequicity. In all of the experiment settings, mT5-small is implemented in the framework and used as the backbone of the framework. We also try to use T5 for monolingual and bilingual settings to analyze the impact of using two different kinds of pre-trained language models (LMs) that will be discussed later in the Subsection \ref{sec:analysis}. We implement reader and evaluator modules for both CamRest and SMD datasets in MinTL, as their paper only included the reader and evaluator for the MultiWOZ~\cite{budzianowski2018multiwoz} dataset. 

%## test on english ##%
\begin{table*}[!t]
\centering
\footnotesize
\begin{tabular}{ l|cccc|cccc }
\toprule
\multirowcell{2}{\textbf{Baselines}} & \multicolumn{4}{c|}{\textbf{CamRest}} & \multicolumn{4}{c}{\textbf{SMD}} \\ 
\cmidrule{2-9} & \textbf{BLEU} & \textbf{Match} & \textbf{Success} & \textbf{Combined} & \textbf{BLEU} & \textbf{Match} & \textbf{Success} & \textbf{Combined} \\ \cmidrule{1-9}
\multicolumn{9}{c}{\texttt{monolingual}} \\ \cmidrule{1-9}
Sequicity \cite{lei2018sequicity} &\textbf{21.17}	&90.30	&80.99	&106.81	&\textbf{21.04}	&82.12	&78.21	&\textbf{101.20} \\
LABES \cite{zhang2020probabilistic} &19.98	&93.64	&72.93	&103.26	&1.94	&80.81	&60.07	&72.39 \\
MinTL \cite{lin2020mintl} &20.34	&83.21	&83.64	&103.77	&18.53	&\textbf{83.90}	&\textbf{78.60}	&99.78 \\
GALAXY \cite{He2021GALAXYAG} &20.53	&\textbf{96.24}	&\textbf{84.62}	&\textbf{110.96}	&18.25	&80.00	&77.32	&96.91\\
\cmidrule{1-9}
\multicolumn{9}{c}{\texttt{bilingual}} \\ \cmidrule{1-9}
Sequicity \cite{lei2018sequicity} &22.53	&91.04	&82.06	&109.08	&21.04	&75.56	&\textbf{83.48}	&100.56 \\
LABES \cite{zhang2020probabilistic} &22.58	&\textbf{97.27}	&79.33	&\textbf{110.88}	&\textbf{22.35}	&\textbf{83.83}	&74.02	&\textbf{101.28} \\
MinTL \cite{lin2020mintl} &\textbf{22.88}	&86.26	&82.26	&107.14	&18.39	&80.23	&79.10	&98.05 \\
GALAXY \cite{He2021GALAXYAG} &20.31	&95.49	&\textbf{84.67}	&110.39	&16.58	&77.37	&76.84	&93.68 \\
\bottomrule
\end{tabular}
\caption{Experiment settings result on English test set. \textbf{bold} denotes the best score per metric.}
\label{tab:engresults}
\end{table*}

\subsection{Hyper-parameters}
We conduct the learning rate search [1e-4, 6e-4, 1e-5] to optimize the performance of MinTL on the new datasets, and the best learning rate is 6e-4. For other ToD frameworks, we adopt their proposed learning rate in our experiments. We use a learning rate of 3e-3 for Sequicity, 1e-4 for GALAXY, and, for LABES, we set the learning rate to 3e-4 and 1e-4 in the CamRest and SMD experiments, respectively. We follow the inference sampling strategies recommended by the original papers that introduce the frameworks. We use a beam size of 5 for generating inference responses in GALAXY, while in the other frameworks, we employ greedy search. We run each experiment once per experiment setting for uniformity with an A100 40GB GPU.

\subsection{Evaluation Metrics}
We use four evaluation metrics for end-to-end ToD tasks to evaluate the quality of responses. There are 1) \textbf{BLEU} score \cite{papineni2002bleu} to assess how fluent the generated response is, 2) \textbf{Match} rate \cite{lin2020mintl}, checks whether the ToD system can produce the entity constraints specified by the user, thereby measuring task completion, 3) \textbf{Success} F1 score \cite{lei2018sequicity}, which determines whether the system has successfully provided all of the information that the user has requested, such as the address or phone number, and 4) \textbf{Combined} score \cite{lin2020mintl, mehri2019structured} using the equation \textbf{Combined} = (Match + Success) $\times$ 0.5 + BLEU. We assess the performance of all frameworks using the CamRest and SMD datasets, along with the designated experiment settings.

\section{Results \& Analysis}

\subsection{Results}
Tables~\ref{tab:indoresults} and~\ref{tab:engresults} show the experiment results for Indonesian and English test sets, respectively. We conclude that GALAXY has a decent score in both monolingual settings for CamRest, and Sequicity with fastText word vector has the best model in monolingual settings for SMD. Furthermore, MinTL outperforms other frameworks in zero-shot cross-lingual settings for both datasets. The non-contextual word-embedding frameworks, such as Sequicity and LABES achieve the highest combined scores for bilingual settings. Furthermore, it was observed that GALAXY does not perform as well as other frameworks when dealing with Indonesian, as we denote its results by using '-' in Table~\ref{tab:indoresults}.

\begin{figure*}[!t]
     \centering
     \begin{subfigure}[b]{0.32\textwidth}
        \centering
        \includegraphics[width=\textwidth]{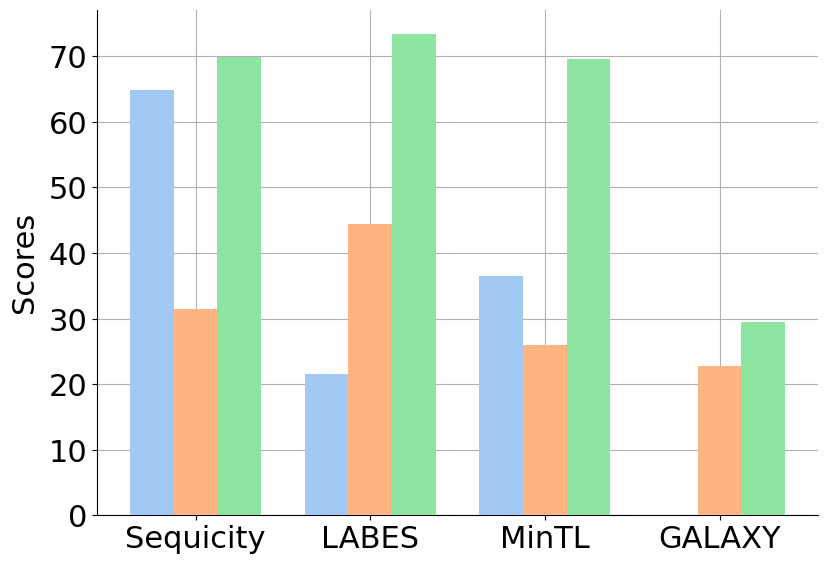}
     \end{subfigure}
     \begin{subfigure}[b]{0.32\textwidth}
        \centering
        \includegraphics[width=\textwidth]{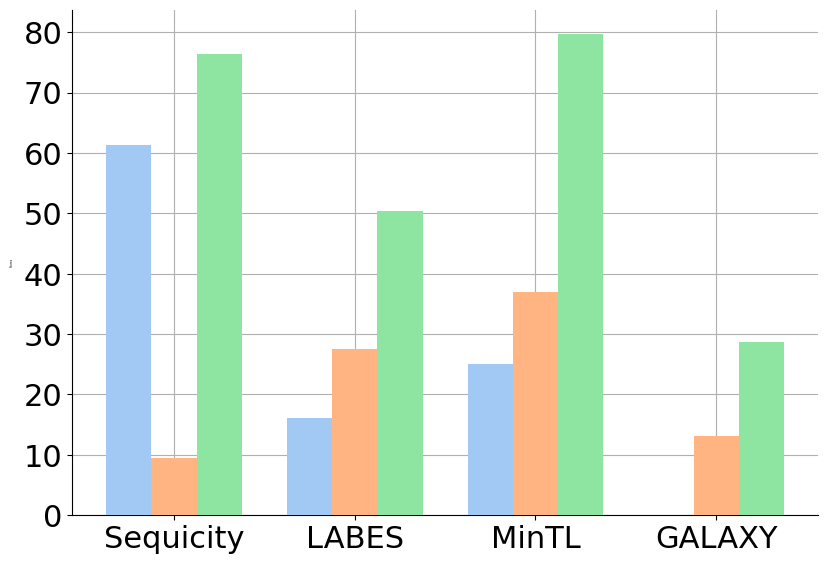}
     \end{subfigure}
     \begin{subfigure}[b]{0.32\textwidth}
        \centering
        \includegraphics[width=\textwidth]{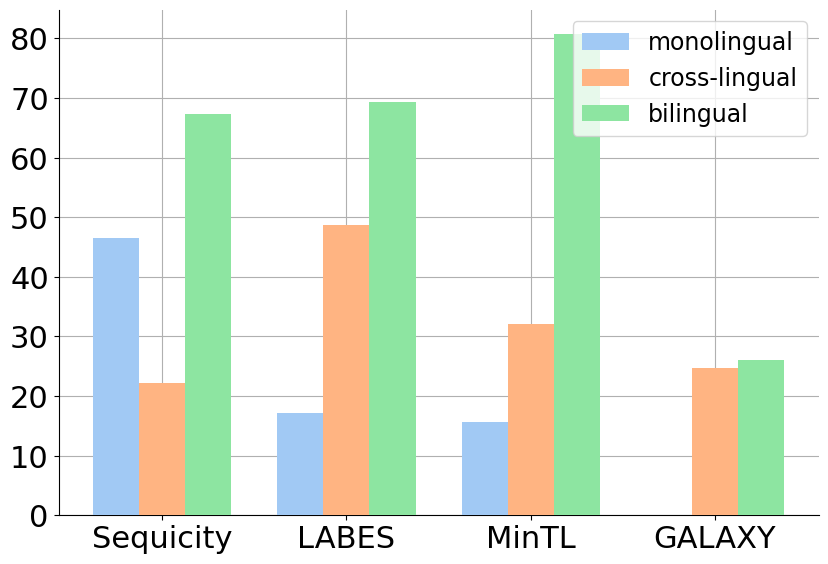}
     \end{subfigure}
     \caption{ToD framework results on IndoSMD per domain. \textbf{Left}: \texttt{POI navigation} domain, \textbf{Center}: \texttt{Calendar scheduling} domain, and \textbf{Right}: \texttt{Weather information} domain.}
     \label{fig:per-domain-result}
\end{figure*}

%ngomongin per domain
Additionally, we present the combined metric results of the ToD frameworks on the IndoSMD test set, categorized by domain, in Figure~\ref{fig:per-domain-result}. We discover that \texttt{POI navigation} is relatively easier than other domains to learn, while \texttt{calendar scheduling} is the hardest domain for ToD frameworks on SMD dataset. Fine-grained results can be found in Appendix Table~\ref{tab:indosmdresultperdomain}.

\subsection{Analysis}
\label{sec:analysis}

% coba bahas tiap model, mungkin bisa suggest kl embedding representation di B Indo ga sebagus di B Inggris
\paragraph{Monolingual English vs. Indonesian Performance.} 
% The monolingual setting results indicate that the experiment metric result is better when we use English corpus compared with Indonesian, as it generally produces lower scores than English. 
In general, the monolingual model trained and tested in English performs better than in Indonesian across different frameworks and datasets. We hypothesize that the pre-trained LMs used in the experiments are better trained to handle English data. GALAXY outperforms the monolingual setting on both Indonesian and English CamRest datasets. However, we find that the GALAXY framework fails to adapt to the IndoSMD dataset and the model is not converged. 
% Regarding scores in the SMD dataset, the lower scores of the Indonesian experiment might be due to a smaller amount of utterances IndoSMD had compared with the original English SMD dataset, which is ten times smaller.
% intinya bilang kalau
We further observe that the experiments using GALAXY in Indonesian are not as effective as English, as evidenced by the superior performance on the English CamRest dataset, despite having the same amount of dialogue and utterance as IndoCamRest. Furthermore, it is worth noting on the IndoToD benchmark that we are not using Indonesian T5 models. This is because there is no paper working on releasing the Indonesian T5 models that are comparable to the English T5 models we use in the experiment.

% Furthermore, it is worth noting that the choice of the backbone model may influence the metric scores in evaluating the MinTL framework. This is due to using the mT5-small model as the backbone of MinTL. It should be acknowledged that, as of the implementation of this experiment, there is no well-documented T5 model specifically trained for Indonesian, so we cannot conduct some experiments using Indonesian-only embedding representation.

% \begin{figure}[!t]
%     \centering
%     \begin{subfigure}{0.95\linewidth}
% 		\centering
% 		\includegraphics[width=\linewidth]{domain-navigate.png}   
% 		\caption{*CL}
% 	\end{subfigure}
% 	\begin{subfigure}{0.95\linewidth}
% 		\centering
%         \includegraphics[width=\linewidth]{domain-schedule.png} 
%         \caption{ISCA}
% 	\end{subfigure}
%     \begin{subfigure}{0.95\linewidth}
% 		\centering
%         \includegraphics[width=\linewidth]{domain-weather.png} 
%         \caption{ISCA}
% 	\end{subfigure}
% 	\caption{Methods used for code-mixing NLP.} % over the years
% 	\label{fig:methods}
% \end{figure}

\paragraph{Transferability from English to Indonesian.} Based on the cross-lingual results, it can be concluded that existing ToD frameworks are unable to handle the task when trained in a different language. While not all training data needs to be in the same language as the test set, it is still essential for the framework to have some knowledge of the target language in order to understand the task at hand. Based on the inference results, we observe that all of the ToD frameworks generate responses in English despite Indonesian user inputs. This can be correlated to the fact that the training data used in the cross-lingual setting is in English. Among all the ToD frameworks that are evaluated in this setting, MinTL performs the best in the cross-lingual setting (due to mT5 as its pre-trained model, trained on multilingual datasets), while GALAXY and Sequicity do not perform well.

\paragraph{Monolingual vs. Bilingual.} \label{sec:monovsbi}
The results demonstrate that, in most cases, the bilingual setting yields higher scores than the monolingual (i.e., Indonesian and English monolingual setting), especially in the Indonesian test set. The bilingual setting has the advantage of using a larger amount of training data and performing tasks in both languages \cite{lin2021bitod}, increasing both the overall metric scores and the individual metric components. We also compare and analyze the framework's performance closely based on the test set we used. The results show that most of the bilingual settings scores are higher than monolingual settings in both the Indonesian and English test sets showing the effective usage of language data counterpart for domain adaptation. 

However, based on the inference result in Table~\ref{tab:inferenceexample}, we observe that the responses generated between bilingual and monolingual settings are not substantially different in terms of the semantics,
% by the system do not explicitly differentiate between the bilingual and monolingual settings, 
confirmed by the fact that the difference in BLEU scores between these two settings is not significant. However, we see some improvements in belief spans generation in the bilingual setting. ToD frameworks can understand the context from the user better when they are trained with a larger amount of training data, although the training data are a mixture of English and Indonesian.
% although it comes in dual language. 
This highlights the significance of leveraging cross-lingual data, in this case, using English data on top of Indonesian data, in the training process of the ToD system to enhance the model performance and achieve better metric scores in the target language.

\begin{figure}[!t]
\centering
\resizebox{\linewidth}{!}{
\includegraphics{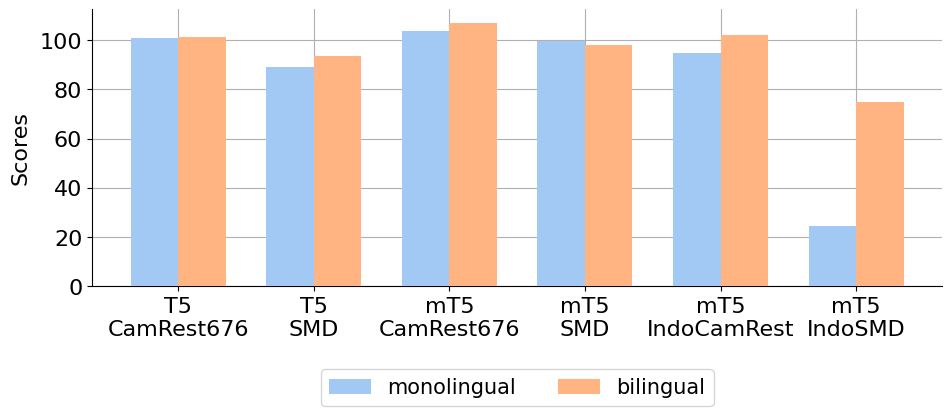}
}
\caption{Monolingual and bilingual MinTL combined score using T5 and mT5 models on CamRest676, SMD, IndoCamRest, and IndoSMD as test sets.}
\label{fig:mintldetailresult}
\end{figure}

Furthermore, we delve deeper into analyzing the potential of bilingual training using the MinTL framework. Interestingly, this framework is flexible since we can modify the pre-trained transformer LM unlike other frameworks. We explore not only English T5, but also multilingual version of T5. We compare the performance of T5-small and mT5-small as the backbone of the MinTL framework, shown in Figure~\ref{fig:mintldetailresult} and Appendix Table~\ref{tab:t5andmt5} for per metric details. Based on our findings, we observe that bilingual training positively impacts the BLEU score, match rate, and combined metrics for the MinTL framework, either using T5 or mT5. The results indicate that the bilingual setting can yield benefits on the efficient training process of T5 and mT5, especially in Indonesian, that is considered as underrepresented language. Moreover, we notice that mT5-small is more effective than T5-small in end-to-end ToD tasks. Overall, the bilingual training is beneficial for training end-to-end ToD, especially to improve the performance on non-English languages.

%## results in bilingual: train bi, test bi ##% siapa tau butuh nanti
% \multicolumn{9}{c}{\textbf{Experiment Setting: \emph{bilingual}}} \\ \cmidrule{1-9}
% Sequicity \cite{lei2018sequicity} &20.05	&93.18	&82.09	&107.69	&19.29	&73.64	&\textbf{79.26}	&95.74 \\
% LABES \cite{zhang2020probabilistic} &\textbf{20.42}	&\textbf{95.45}	&79.52	&107.91	&\textbf{20.26}	&\textbf{77.61}	&73.62	&\textbf{95.88} \\
% MinTL \cite{lin2020mintl} &16.77	&85.94	&83.53	&101.50	&15.83	&76.52	&78.75	&93.46 \\
% GALAXY \cite{He2021GALAXYAG} &18.78	&95.11	&\textbf{84.16}	&\textbf{108.42}	&14.59	&73.44	&70.20	&86.41\\

% contoh error analysis mungkin bisa ditaruh di appendix
\paragraph{Error Analysis.} We conduct an error analysis to investigate the limitations of existing frameworks. In general, most issues occur in the form of a ``template response''. The end-to-end ToD systems sometimes generate responses that appear to be template-like despite varying user inputs
% It usually occurs if the systems receive user inputs that are on the same domain, which appears to produce similar outputs based on what domain this system is handling at that time.
which might be because of the limited trainable parameters and dataset used in the training process.
Despite its goal to complete the user's task, the ToD must also be able to maintain the user's satisfaction whenever using the ToD as a product. Users may become disinterested if they receive repeated responses, potentially leading to reduced usage.
Another issue we notice is the occurrence of repeated tokens in the generated output by the framework during the inference phase. Specifically, we find that the Sequicity and MinTL frameworks tend to produce responses with repeated tokens more than the other ToD frameworks. This could be due to the decoding method applied by these frameworks in our research which relied on a greedy search approach. 
% for some of the frameworks. 
The other issue that is worth mentioning in inference examples is the hallucination in the generated outputs. Some of the outputs do not have the correct combination of slots that are found in the training set.

\begin{table*}
\centering
\resizebox{\linewidth}{!}{
    \begin{tabular}{p{.20\linewidth}|p{.20\linewidth}|p{.60\linewidth}}
    \toprule
    \textbf{Input} & \textbf{Belief spans} & \textbf{Response} \\ \midrule
    "carikan saya restoran mahal yang menyediakan makanan internasional."  \newline(\textit{find me an expensive restaurant that serves international food}) & \textbf{gold:} {"restaurant-food": "internasional", "restaurant-pricerange": "mahal"} \newline 
\textbf{monolingual:} {"restaurant-food": "internasional", "restaurant-pricerange": "menengah"} \newline \textbf{bilingual:} {"restaurant-food": "internasional", "restaurant-pricerange": "mahal"} & \textbf{gold:} "maafkan saya. saya tidak bisa menemukan restoran apapun yang menyediakan makanan internasional di kisaran harga yang mahal" \newline (\textit{I am sorry. I couldn't find any restaurant that provides international food at such an expensive price range}) \newline \textbf{monolingual:} "value\_name adalah restoran mahal yang menyediakan makanan internasional. apakah kamu ingin tahu lokasinya?" \newline (\textit{value\_name is an expensive restaurant that serves international food. Would you like to know its location?}) \newline \textbf{bilingual:} "value\_name menyediakan makanan internasional. apakah kamu ingin informasi lebih lanjut?" \newline (\textit{value\_name serves international foods. Would you like other further information?}) \\
    \bottomrule
    \end{tabular}
}
\caption{Example of the LABES response on IndoCamRest using monolingual and bilingual settings. The English translation is in \textit{italic}.}
\label{tab:inferenceexample}
\end{table*}

\section{Related Work}

% Jelasin dulu yg TODS bahasa inggris, baru specific ke indo - ka genta

% content:

% 1. Sebutin & jelasin dikit ttg datasets e2e ToD yang tersedia (English, bilingual). (terus bilang kalau belum ada yg Indonesian (1 kalimat aja)) --> kayaknya better kalau paragraf ini yang paling panjang di section in (masih bingung kira2 mau gimana manjanginnya)
\paragraph{ToD Datasets.} Multiple studies have contributed to creating datasets for ToD using various approaches. bAbI \cite{bordes2016learning} is an English synthetic dataset combining five subtasks into ToD system datasets. SMD \cite{eric2017key} uses the WoZ scheme to build an English car assistant dialogue system that has knowledge of weather and places. Other English datasets using the WoZ scheme are CamRest \cite{wen2017network}, a dataset for a restaurant reservation, MultiWoZ \cite{budzianowski2018multiwoz}, a multi-domain WoZ dataset containing five domains collected using the human-to-human scheme, and OpenDialKG \cite{moon2019opendialkg}, a synthetic dataset containing four domains. On non-English datasets. such as Chinese, there are CrossWoZ \cite{zhu2020crosswoz}, a large-scale Chinese WoZ dataset with five domains, and RiSAWoZ~\cite{quan2020risawoz}, a large-scale multidomain Chinese WoZ dataset with rich semantic annotations up to 12 domains. MetaLWOz \cite{lee2019multi-domain} is a goal-oriented dialogue corpus containing 51 domains and collected using WoZ.
Other datasets include bilingual languages, such as BiToD \cite{lin2021bitod}, a dataset for tourism assistants focusing on English and Chinese languages, and COD~\cite{majewska2023cross}, a dataset containing Russian, Arabic, Indonesian, and Swahili. 
% tambahin CrossWOZ, RiSAWOZ
% XPersona \cite{lin2021xpersona}, a multilingual chitchat dataset of persona conversations in six languages besides English for evaluating multilingual personalized agents. 
Based on the aforementioned datasets, we conclude that the ToD dataset with contextual knowledge focusing on Indonesian is not yet available.

\paragraph{Indonesian Dialogue Systems Datasets.} Research for the ToD system in the Indonesian language is still underway. IndoNLG \cite{cahyawijaya2021indonlg}, a benchmark for natural language generation in low-resource languages, focuses on summarization, question answering, chit-chat, and machine translation for Indonesian, Javanese, and Sundanese languages. An Indonesian language subset of the XPersona \cite{lin2021xpersona} dataset, which consists of approximately 17 thousand dialogues, is used for the chit-chat task. It is worth noting that although IndoNLG is making significant progress, it has not yet explored the ToD system. Furthermore, NusaCrowd~\cite{cahyawijaya2022nusacrowd}, a large pool of hundreds of Indonesian NLP data, only listed one Indonesian ToD dataset, i.e.,  COD~\cite{majewska2023cross}, which is a multilingual ToD dataset that is solely used for evaluation with only 194 dialogues spanning across 11 domains. Therefore, the Indonesian ToD dataset is still very limited, and urgent action is required to increase the coverage of Indonesian ToD datasets.

\paragraph{Frameworks for End-to-End ToD.} 
% RewardNet \cite{feng2023fantastic} is a framework that uses reward-function learning, inspired by the classical learning-to-rank literature to guide the training of the end-to-end ToD agent. 
GALAXY \cite{He2021GALAXYAG} is a pre-trained dialogue model that uses semi-supervised learning to learn dialogue policy from limited labeled dialogues and large-scale unlabeled dialogue corpora. LABES \cite{zhang2020probabilistic}, conversely, represents belief states as discrete latent variables and models them jointly with system responses given user inputs. MinTL \cite{lin2020mintl} efficiently uses pre-trained LMs in developing task-oriented dialogue systems, eliminating the need for ad hoc modules in studies that use this framework. DAMD \cite{zhang2020task} is a network that accommodates state-action pair structure in generation and can leverage the proposed multi-action data augmentation framework to address the multidomain response generation problem. Sequicity \cite{lei2018sequicity} uses the Seq2Seq model for dialogue state tracking and generating a response to the user. Some other recent works~\cite{bang2023multitask,hudeček2023llms} explore the potential of using large LMs for performing zero-shot ToD. In this paper, we utilize GALAXY, LABES, MinTL, and Sequicity to evaluate the performance of various end-to-end ToD frameworks.

%% List of resources: 6 e2e frameworks: Sequicity, LABES, MinTL, GALAXY, DAMD & RewardNet yg ga dipake. 8 e2e datasets: english (bAbI, SMD, CamRest676, MultiWOZ, OpenDialKG), bilingual (BiToD, COD (3 languages)), multilingual-chitchat (XPersona). IndoNLG
%% link to resources: https://drive.google.com/drive/folders/19I5vb6M78pMml2Bsi9MaVz36ei0YprFf?usp=sharing

\section{Conclusion}
We introduce IndoToD, an end-to-end multi-domain Indonesian ToD benchmark. We extend the existing two English ToD datasets using an efficient framework by delexicalizing the conversation into templates and reconstruction using KB. We evaluate our benchmark in monolingual, cross-lingual, and bilingual settings. We also show the benefits of having English data to improve Indonesian performance. 
% using our proposed dataset building pipelines and the execution of suggested experiment settings: monolingual, cross-lingual, and bilingual, to evaluate and analyze existing ToD system frameworks' performance. The experiment result includes the metric scores of several experiment settings in each Indonesian and English test set, which shows the potential benefits of a bilingual (Indonesian-English) approach in developing an end-to-end ToD system. 
Furthermore, we conduct some analysis from multiple viewpoints that shows the existing ToD system's inability to understand dialogue context from unseen language and the current ToD system's behavior in handling underrepresented language such as Indonesian using our datasets. We also investigate some errors that usually occur in ToD system responses to explore its limitation.
% There are several directions for this future work. First, we can evaluate more ToD system frameworks by analyzing their performance using our experiment settings and investigating various parameters i.e. using T5-base/large instead of small for MinTL framework, adjusting the learning rate, and exploring alternative decoding methods. This approach will contribute to the development of a more comprehensive benchmark. Second, the human evaluation methods can be conducted to get better analysis and represent ToD frameworks' performance better \cite{liu2016not}. Despite its limitations, we hope that our research can serve as a high-quality benchmark for other researchers in developing Indonesian end-to-end ToD systems by publicly releasing the datasets, aiming to facilitate and support future advancements in the field.

\section{Limitations}
% size data buat annotation, baseline yg digunakan cm yang publicly available
There are several limitations that hinder our progress in creating the first Indonesian end-to-end ToD benchmark. Firstly, we encountered a limitation in translating the English dataset. Due to the constraints of our resources, specifically having only three annotators, we could only translate the CamRest dataset and a mere 10\% of the SMD dataset. Expanding the dataset size would significantly contribute to advancing the Indonesian ToD system. Second, we only use publicly available end-to-end ToD system framework repositories as our baselines. Based on MultiWoZ benchmark\footnote{\url{https://github.com/budzianowski/multiwoz}}, there are several end-to-end ToD frameworks with slightly superior metric scores compared to our baselines~\cite{budzianowski2018multiwoz}. But because of their unavailability, we could not evaluate them and only focused on evaluating the best available ToD frameworks possible.

\section{Ethics Statement}
Our work spotlights a need to develop ToD systems for underrepresented languages such as Indonesian through our new ToD datasets: IndoCamRest and IndoSMD. During our study, we commit to the ethical principles of NLP research and are well aware of its impact on the language community. We strongly believe that there is no ethical issue within this work, including the data collection process, annotation, existing dataset usage, and experiments. Within our work, annotators are well-rewarded and give us informed consent as they understand and agree to provide their annotations to build new ToD datasets. We uphold our annotator's privacy and follow the data protection and privacy regulation for releasing the datasets. The dataset itself is free from abusive language and personal information as our ultimate goal in this work is to contribute and make an impact on society with provide useful NLP task resources and more linguistic diversity in the NLP field.

% Entries for the entire Anthology, followed by custom entries
% \bibliography{anthology,custom}
\bibliography{custom}
\bibliographystyle{acl_natbib}

\appendix
\setcounter{table}{0}
\setcounter{figure}{0}

\section{Annotator Recruitment}
\label{sec:annotatorrecruitment}
% 2 native speakers Indonesian, bisa bahasa inggris, demography range dari 19-25 tahun, terus kasih tau tentang gaji 22583 per jam (higher than minimum wage per hour in West Java (19038)--> kasih tau brp persennya cite:https://www.bps.go.id/indicator/19/1172/1/upah-rata---rata-per-jam-pekerja-menurut-provinsi.html)
We recruit two Indonesian native speakers that can speak English to translate the delexicalized English utterance template to Indonesian. The age of the annotators ranged from 19 to 25, and they were college students by profession. They work around 18 hours in total, and they are compensated approximately 22,583 IDR per hour for their work, 18,62\% bigger than the worker minimum wage in West Java (19,038 IDR per hour)\footnote{\url{https://www.bps.go.id/indicator/19/1172/1/upah-rata---rata-per-jam-pekerja-menurut-provinsi.html}}.

\section{Sample Evaluation Correction by Annotators}
\label{sec:samplecorrection}
Appendix Table~\ref{tab:sampledevaluationfixing} shows several examples of evaluation by annotators. The evaluation process is intended to maintain high-quality translation from the annotator. We report five sampled English utterances, incorrect Indonesian utterances, and corrected Indonesian utterances by annotators.

\section{Per Domain Experiment Settings Result on IndoSMD Test Set}
\label{sec:indosmdperdomain}
We report per domain results for all of the baselines on IndoSMD dataset on Appendix Table~\ref{tab:indosmdresultperdomain}. We can conclude that Sequicity has the best score on monolingual Indonesian setting, meanwhile MinTL outperforms other frameworks on cross-lingual and bilingual settings by having the most biggest metric score compared to others when handling Indonesian test set.

\section{MinTL Result Comparisons}
\label{sec:mintlresultcomparisons}
We present MinTL result comparisons by using two pre-trained LM: T5 and mT5 on Appendix Table~\ref{tab:t5andmt5}. The table compares monolingual and bilingual approach on English test set. The result shows that bilingual setting have a beneficial impact on MinTL to learn the task well compares to monolingual setting.

\section{Annotator Instruction and Consent}
\begin{figure*}[!t]
\centering
\resizebox{\linewidth}{!}{
\includegraphics{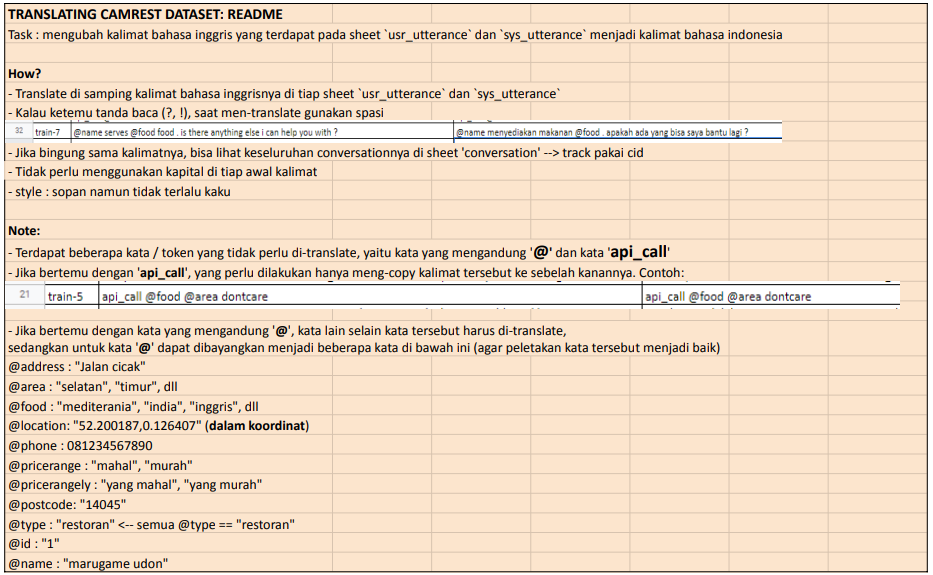}
}
\caption{Instruction for annotators to translating English CamRest templates.}
\label{fig:camrestinstruction}
\end{figure*}

\begin{figure*}[!t]
\centering
\resizebox{\linewidth}{!}{
\includegraphics{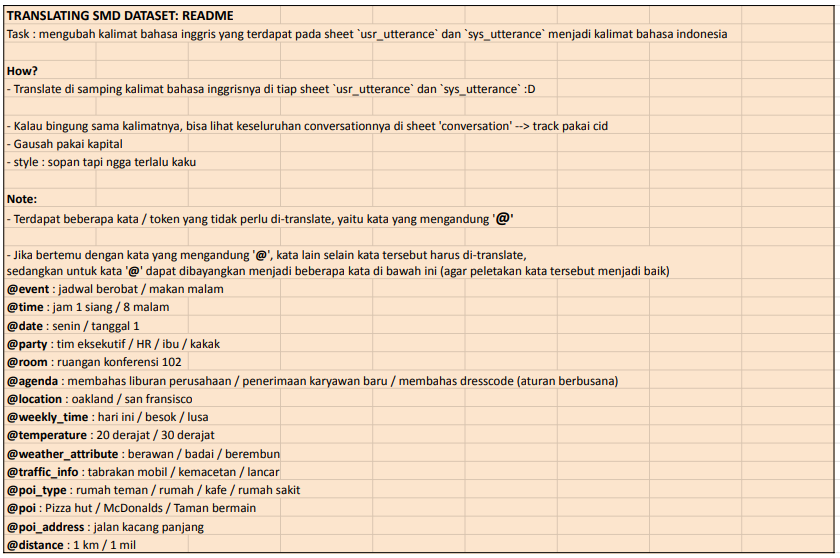}
}
\caption{Instruction for annotators to translating English SMD templates.}
\label{fig:smdinstruction}
\end{figure*}

\begin{figure*}[!t]
\centering
\resizebox{0.80\linewidth}{!}{
\includegraphics{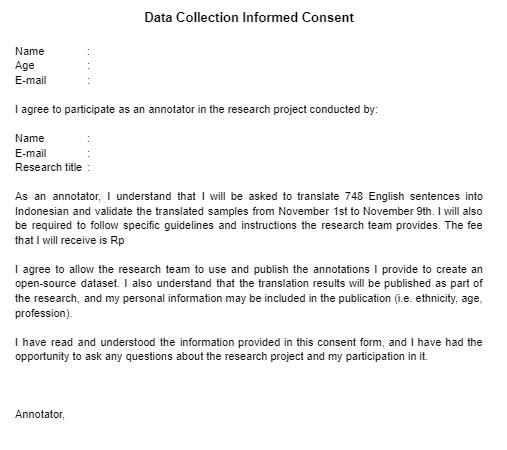}
}
\caption{Annotator informed consent template.}
\label{fig:informedconsent}
\end{figure*}

We report instructions for annotators to translate several English CamRest and SMD utterance templates, as well as annotator informed consent templates on Appendix Figure~\ref{fig:camrestinstruction}, Appendix Figure~\ref{fig:smdinstruction}, and Appendix Figure~\ref{fig:informedconsent}. The instructions are in Indonesian as all of the annotators are Indonesian native speakers.

% buat 2 contoh tabel fix after dataset construction: partikel
% english | indo salah | indo bener
\begin{table*}
\centering
\footnotesize
\resizebox{0.95\linewidth}{!}{
\begin{tabular}{ p{.30\linewidth}|p{.30\linewidth}|p{.30\linewidth} }
\toprule
\textbf{English Utterance} & \textbf{Incorrect Indonesian utterance} & \textbf{Corrected Indonesian utterance} \\ \midrule
"remind me that i have a doctors appointment at 5 pm" & "ingatkan saya bahwa saya akan janji temu dengan dokter pada pukul 5 sore" & "ingatkan saya bahwa saya \textbf{memiliki} janji temu dengan dokter pada pukul 5 sore" \\ \midrule
"that you, that's all I need to know." & "itu kamu, hanya itu saja yang perlu saya ketahui" & "\textbf{terimakasih}, hanya itu saja yang perlu saya ketahui" (there is a typo on the English utterance. It suppose to be 'thank you', not 'that you') \\ \midrule
"are there ant jamaican restaurants in any part of town?" & "apakah ada restoran semut jamaika di bagian manapun dari kota ini?" & "apakah ada restoran jamaika di bagian manapun dari kota ini?" (there is a typo on the English utterance. It suppose to be 'any', not 'ant') \\ \midrule
"la tasca is located at 14-16 Bridge Street and the postcode is C.B 2, 1 U.F.  Their phone number is 1223 464630. May I help you with anything else?" & "la tasca berlokasi di 14 -16 bridge street dan kode pos nya adalah c.b 2, 1 u.f. nomornya adalah 1223 464630. bisakah saya membantu kamu dengan hal yang lain?" & "la tasca berlokasi di 14 -16 bridge street dan kode pos nya adalah c.b 2, 1 u.f. nomornya adalah \textbf{01223 464630}. bisakah saya membantu kamu dengan hal yang lain?" (we check in to the KB and we find out the correct entity for its phone number is '01223 464630') \\ \midrule
"it is located at  Quayside Off Bridge street.  The phone number is is 01223 301030." & "alamatnya adalah quayside off bridge street. nomor teleponnya adalah 01223 301030." & "alamatnya adalah quayside off bridge street. nomor teleponnya adalah \textbf{01223 307030}." (we check in to the KB and find out that the correct entity is 01223 307030) \\
\bottomrule
\end{tabular}
}
\caption{Sample evaluation correction by annotators.}
\label{tab:sampledevaluationfixing}
\end{table*}

%## smd per domain test split result ##%
\begin{table*}
\centering
\footnotesize
\resizebox{0.95\linewidth}{!}{
\begin{tabular}{ l|cccc|cccc|cccc }
\toprule
\multirowcell{3}{\textbf{Baselines}} & \multicolumn{12}{c}{\textbf{Domain}} \\ 
\cmidrule{2-13} & \multicolumn{4}{c|}{\textbf{POI Navigation}} & \multicolumn{4}{c}{\textbf{Calendar Scheduling}} & \multicolumn{4}{|c}{\textbf{Weather Information}} \\ 
\cmidrule{2-13} & \textbf{BL} & \textbf{Ma} & \textbf{Su} & \textbf{Co} & \textbf{BL} & \textbf{Ma} & \textbf{Su} & \textbf{Co} & \textbf{BL} & \textbf{Ma} & \textbf{Su} & \textbf{Co} \\ \cmidrule{1-13}
\multicolumn{13}{c}{\texttt{monolingual}} \\ \cmidrule{1-13}
Sequicity &\textbf{7.90}	&25.00	&\textbf{89.05}	&\textbf{64.93}
&\textbf{6.85}	&\textbf{61.90}	&\textbf{46.91}	&\textbf{61.26}
&\textbf{10.57}	&\textbf{33.33}	&\textbf{38.71}	&\textbf{46.59} \\
LABES &1.30	&0.00	&40.50	&21.55
&0.00	&0.00	&32.30	&16.15
&0.00	&14.30	&20.00	&17.15 \\
MinTL &0.02	&\textbf{50.00}	&22.92	&36.48
&0.34	&19.05	&30.30	&25.02
&0.004	&4.76	&26.51	&15.64 \\
GALAXY &- &- &- &-
&- &- &- &-
&- &- &- &-\\ \cmidrule{1-13}
\multicolumn{13}{c}{\texttt{cross-lingual}} \\ \cmidrule{1-13}
Sequicity &0.00	&12.50	&50.53	&31.52
&0.00	&14.29	&4.82	&9.56
&0.00	&19.05	&25.45	&22.25 \\
LABES &0.00	&13.30	&\textbf{75.50}	&\textbf{44.40}
&0.00	&\textbf{16.70}	&38.30	&27.50
&0.00	&\textbf{42.90}	&54.50	&\textbf{48.70} \\
MinTL &0.00	&\textbf{16.67}	&35.29	&25.98
&\textbf{0.004}	&14.29	&\textbf{59.54}	&\textbf{36.92}
&0.00	&4.76	&\textbf{59.41}	&32.09 \\
GALAXY &\textbf{0.01}	&11.76	&33.66	&22.72
&0.00	&0.00	&26.17	&13.09
&0.00	&0.00	&49.54	&24.77\\ \cmidrule{1-13}
\multicolumn{13}{c}{\texttt{bilingual}} \\ \cmidrule{1-13}
Sequicity &\textbf{8.92}	&44.44	&77.47	&69.88
&11.73	&\textbf{61.90}	&67.37	&76.37
&11.72	&\textbf{57.14}	&54.05	&67.32 \\
LABES &4.50	&\textbf{60.00}	&\textbf{77.90}	&\textbf{73.45}
&6.90	&33.30	&53.70	&50.40
&9.70	&42.90	&76.30	&69.30 \\
MinTL &4.75	&55.56	&74.24	&69.65
&\textbf{15.94}	&47.62	&\textbf{80.00}	&\textbf{79.75}
&\textbf{13.85}	&52.38	&\textbf{81.54}	&\textbf{80.81} \\
GALAXY &0.06	&47.06	&11.76	&29.47
&0.06	&50.00	&7.23	&28.68
&0.12	&42.86	&8.96	&26.03 \\ 
\bottomrule
\end{tabular}
}
\caption{Per domain experiment settings result on Indonesian test set using SMD dataset. \textbf{bold} denotes the best score per metric. BL, Ma, Su, and Co denote the \textbf{BL}EU score, \textbf{Ma}tch rate, \textbf{Su}ccess F1, and \textbf{Co}mbined score respectively.}
\label{tab:indosmdresultperdomain}
\end{table*}

%% t5 and mt5 %%
\begin{table*}
\centering
\footnotesize
\resizebox{0.95\linewidth}{!}{
\begin{tabular}{ r|cccc|cccc }
\toprule
\multirowcell{2}{\textbf{Experiment Setting}} & \multicolumn{4}{c|}{\textbf{CamRest}} & \multicolumn{4}{c}{\textbf{SMD}} \\ 
\cmidrule{2-9} & \textbf{BLEU} & \textbf{Match} & \textbf{Success} & \textbf{Combined} & \textbf{BLEU} & \textbf{Match} & \textbf{Success} & \textbf{Combined} \\ \cmidrule{1-9}
\multicolumn{9}{c}{\texttt{T5-small}} \\ \cmidrule{1-9}
Monolingual &17.57	&83.21	&\textbf{83.25}	&100.80 
&15.31	&71.54	&76.22	&89.19\\
Bilingual &\textbf{18.50}	&\textbf{85.50}	&80.35	&\textbf{101.42}
&\textbf{16.23}	&\textbf{75.29}	&\textbf{79.52}	&\textbf{93.63}\\
\cmidrule{1-9}
\multicolumn{9}{c}{\texttt{mT5-small}} \\ \cmidrule{1-9}
Monolingual &20.34	&83.21	&\textbf{83.64}	&103.77
&\textbf{18.53}	&\textbf{83.90}	&78.60	&\textbf{99.78}\\
Bilingual &\textbf{22.88}	&\textbf{86.26}	&82.26	&\textbf{107.14}
&18.39	&80.23	&\textbf{79.10}	&98.05\\
\bottomrule
\end{tabular}
}
\caption{MinTL \cite{lin2020mintl} result comparisons using pre-trained LM: T5 and mT5 as backbone model on English test set. \textbf{bold} denotes the best score per metric.}
\label{tab:t5andmt5}
\end{table*}

\end{document}